\documentclass{article}

\usepackage{microtype}
\usepackage{graphicx}
\usepackage{subcaption}
\usepackage{amsmath}
\usepackage{amsfonts}
\usepackage{bm}
\usepackage{booktabs} %

\DeclareMathOperator*{\argmax}{arg\,max}
\DeclareMathOperator*{\argmin}{arg\,min}

\usepackage{hyperref}

\usepackage[accepted]{icml2019}

\icmltitlerunning{Bayesian Modelling in Practice: Using Uncertainty to Improve Trustworthiness in Medical Applications}

\begin{document}

\twocolumn[
\icmltitle{Bayesian Modelling in Practice: Using Uncertainty to Improve Trustworthiness in Medical Applications}

\begin{icmlauthorlist}
\icmlauthor{David Ruhe}{uva,pm}
\icmlauthor{Giovanni Cin\`{a}}{pm}
\icmlauthor{Michele Tonutti}{pm}
\icmlauthor{Daan de Bruin}{pm}
\icmlauthor{Paul Elbers}{amc}
\end{icmlauthorlist}

\icmlaffiliation{uva}{University of Amsterdam}
\icmlaffiliation{pm}{Pacmed BV, Amsterdam}
\icmlaffiliation{amc}{Univeristy Medical Center, Amsterdam}

\icmlcorrespondingauthor{David Ruhe}{david.ruhe@gmail.com}

\icmlkeywords{Machine Learning, ICML}

\vskip 0.3in
]

\printAffiliationsAndNotice{}  %

\begin{abstract}
The Intensive Care Unit (ICU) is a hospital department where machine learning has the potential to provide valuable assistance in clinical decision making. 
Classical machine learning models usually only provide point-estimates and no uncertainty of predictions. 
In practice, uncertain predictions should be presented to doctors with extra care in order to prevent potentially catastrophic treatment decisions. 
In this work we show how Bayesian modelling and the predictive uncertainty that it provides can be used to mitigate risk of misguided prediction and to detect out-of-domain examples in a medical setting. 
We derive analytically a bound on the prediction loss with respect to predictive uncertainty. 
The bound shows that uncertainty can mitigate loss. 
Furthermore, we apply a Bayesian Neural Network to the MIMIC-III dataset, predicting risk of mortality of ICU patients. 
Our empirical results show that uncertainty can indeed prevent potential errors and reliably identifies out-of-domain patients. These results suggest that Bayesian predictive uncertainty can greatly improve trustworthiness of machine learning models in high-risk settings such as the ICU.
\end{abstract}

\section{Introduction}
The Intensive Care Unit (ICU) is a resource-intensive environment where patients receive care that is not readily available elsewhere in the hospital. 
ICU patients usually are in life-threatening conditions. 
Therefore, adequate assessment of illness severity and expected effectiveness of interventions is of utmost importance for clinical decision making. 
Recent advancements in machine learning have cleared the path to deployment of machine learning based decision support software systems in critical areas like the ICU. 
Examples are predicting risk of readmission after discharge \citep{thoral2018right}, optimizing sepsis treatment \citep{raghu2017deep}, or predicting mortality risk given the current state of the patient \citep{pirracchio2015mortality}.
While these applications have proven to be effective, model trustworthiness remains elusive. 
Most machine learning models only provide point estimates of their parameters and corresponding predictions. 
A recent study has shown that these models can output high-risk predictions on datapoints that lie far from their observed dataset \citep{nguyen2015deep}.
Especially in a critical area like the ICU these flaws could result in catastrophes. 
For effective machine learning based decision support, we think it is crucial that machine learning methods output uncertainty estimates beside predictions. 
Bayesian modelling has the desirable property of expressing predictive uncertainty through stochasticity in its parameters.
In this work, we show how \textit{Bayesian Neural Networks}, through predictive uncertainty (hereafter referred to as uncertainty), can correctly identify predictions that are likely to be misguided. 
When the model encounters a datapoint that lies far from its observed set, a practitioner can be notified through the model's uncertainty and observe that the patient has a combination of symptoms that hitherto has not been seen. 
Therefore, the practitioner should proceed with care.

In previous work, \citet{leibig2017leveraging} are among the first to apply model uncertainty to healthcare by diagnosing diabetic retinopathy from fundus images. 
Similarly, \citet{nair2018exploring, wang2019aleatoric, orlando2019u2} also estimate uncertainty to image analysis using MC Dropout \cite{gal2016uncertainty}.
We extend on previous work by making the following constributions.
\begin{enumerate}
    \item We provide mathematical bounds on the obtainable loss with respect to uncertainty.
    \item Through these bounds, model performance is directly related to uncertainty. Uncertainty prevents high-loss prediction errors.
    \item We show that BNNs can identify out-of-domain patients competently in a real use-case.
\end{enumerate}
Additionally, we are (to our best knowledge) novel in the approach of using Bayes By Backprop \cite{blundell2015weight} instead of MC Dropout to estimate model distribution parameters directly.
This choice has two motivations.
First, MC Dropout rates have to be carefully adjusted to obtain well-calibrated uncertainties. 
This requires tuning all dropout probabilities, which is unfeasible for deep neural networks.
Second, the MC dropout approximate posterior does not contract with more data, and therefore the approach has been questioned \cite{osband2016risk}.
We also extend on earlier publications by applying Bayesian uncertainty to (ICU) signal processing.

The rest of the paper is structured as follows. 
In section \ref{sec: background} we discourse required background knowledge. 
In section \ref{sec: methods} we elaborate on methodological decisions for signal processing and modelling. 
In section \ref{sec: results} we provide an overview of the results, showing that model uncertainty can mitigate the prediction loss and how uncertainty relates to out-of-domain observations. 
In section \ref{sec: conclusion} we conclude and provide suggestions for future research directions.

\section{Background}
\label{sec: background}
In this section we summarize important background theory about Bayesian modelling.
\subsection{Bayesian Neural Networks}
In deterministic modelling, given a dataset of observations $\bm{X} = \{ \bm{x_1}, \dots, \bm{x_N}\}$ and labels $\bm{y} = \{y_1, \dots, y_N\}$, we restrict model parameters $\bm{\omega}$ to a point estimate and optimize the likelihood function $p(\bm{y|X}, \bm{\omega})$ directly.
That is, we try to find 
\begin{equation}
\label{eq:mll}
\bm{w^*} = \argmax_{\bm{\omega}} p(\bm{y|X}, \bm{\omega}).
\end{equation}
To capture model uncertainty, we strive to find the posterior distribution over the model parameters:
\begin{equation}
p(\bm \omega | \bm X, \bm y) = \frac{p(\bm y | \bm X, \bm \omega) p(\bm \omega)}{p(\bm y | \bm X)},
\end{equation}
where $p(\bm{\omega})$ is a prior. 
The marginal likelihood $p(\bm y | \bm X)$ is computed by: $\int d {\bm \omega} p(\bm y | \bm X, \bm \omega)p(\bm \omega)$. 
Given many continuous model parameters this term cannot be evaluated analytically, which is the case when $\bm \omega$ are the parameters of a (deep) neural network.
Consequently, the posterior is intractable as well.
Therefore, we approximate it with a tractable variational distribution $q_{\bm \theta}$ through a procedure called \textit{variational inference} \citep{hinton1993keeping}.
The goal is to minimize the KL divergence between the true and approximate posterior:
\begin{align}
\label{eq:dkl}
\bm {\theta^*} &= \argmin_{\bm \theta} \text{KL}\left(q_{\bm \theta}(\bm \omega | \bm X, \bm y) || p(\bm \omega | \bm X, \bm y)\right). %
\end{align}
This KL divergence, too, is analytically intractable.
Instead, we can minimize it by maximizing the \textit{evidence lower bound}:
\begin{align}
\label{eq:elbo}
\mathcal{L}_{VI}(\bm \theta) &{:=} \mathbb{E}_q \left[ \log p(\bm y | \bm X, \bm \omega)  \right] - \text{KL}(q_{\bm \theta}(\bm \omega | \bm X, \bm y) || p (\bm \omega)) \\ 
&= \log p(\bm{y|X}) - \text{KL}\left(q_{\bm \theta}(\bm \omega | \bm X, \bm y) || p(\bm \omega | \bm X, \bm y)\right)\nonumber
\end{align}
Since the KL-divergence is non-negative, the evidence lower bound lower bounds the marginal likelihood. 
Thus, we are able to minimize the gap between the true and approximate posterior.

In this paper, we used the approach of \citet{blundell2015weight}, coined Bayes By Backprop (BBB), to maximize the evidence lower bound.
Predictions for a data point $\bm x_i$ are made by sampling model parameters from the variational posterior distribution and taking the mean of the predictions of all sampled models.
That is: 
\begin{equation}
    p(y_i | \bm x_i) = \frac{1}{T} \sum_{t=1}^T p(y_{i}| f^{\bm {\omega}_t} (\bm x_i))
\end{equation}
with $\bm{\omega}_t \sim q^*_\theta(\bm{\omega})$.
Predictive uncertainty was computed as the variance in the predictions: 
\begin{equation}
Var(y_i|\bm x_i) =  \frac{1}{T} \sum_{t=1}^T (p(y_i|\bm x_i) - p(y_i|f^{{\bm \omega}_t}(\bm x_i))^2
\end{equation}

\section{Methods}
\label{sec: methods}
\subsection{Data and Preprocessing}
Data was obtained from the Multiparameter Intelligent Monitoring in Intensive Care
(MIMIC-III v1.4) database \citep{johnson2016mimic}.
This dataset contains signal data for 46,520 patients and 58,976 ICU admissions.
$7,821$ of the patients in the MIMIC-III dataset are newborns; these were excluded from the dataset as newborns come with different characteristics and treatment requirements.
For each patient, features were constructed by aggregating relevant clinical information, lab values and vital signs. 
The objective of the model was to find patients that have a high risk of mortality during the admission.
More information on the task and data processing is included in the supplementary material.

\subsection{Model Architecture}
Following \citet{blundell2015weight}, we used a Gaussian scale mixture as our prior.
We used two 128-neuron hidden layers, ReLU intermediate activation and Sigmoid final activation to obtain the probability distribution $p(\bm y_i | f^{\omega}(\bm{x_i}))$. 
All weight distributions were initialized with $\bm \mu \sim \mathcal{U}(-.2, .2)$ and $\bm \rho \sim \mathcal{U}(-5, -4)$. 
Adam was used as the optimizer with the configurations as suggested by the authors \citep{kingma2014adam}.\footnote{Code is available at \url{https://github.com/Pacmed/aisg_2019}}

\section{Results}
\label{sec: results}
We discuss our results in the following two subsections.

\subsection{Uncertainty Mitigates Prediction Loss}
\begin{figure}
    \centering
    \includegraphics[width=0.6\linewidth]{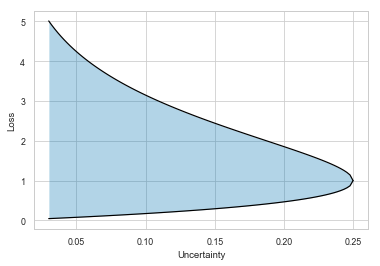}
    \caption{The reachable loss region in a Bayesian binary classification problem.}
    \label{fig:loss_uncertainty}
\end{figure}

\begin{figure}
    \centering
    \begin{subfigure}{.6\linewidth}
        \centering
        \includegraphics[width=\linewidth]{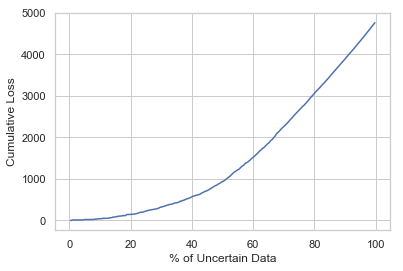}
        \caption{}
    \end{subfigure}
    \begin{subfigure}{.6\linewidth}
        \centering
        \includegraphics[width=\linewidth]{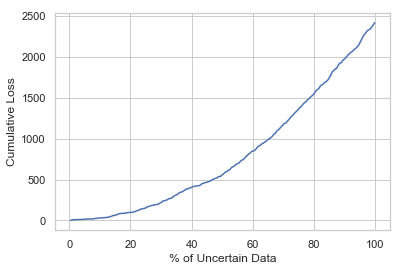}
        \caption{}
        \label{fig: perf_std_gb}
    \end{subfigure}
\caption{Cumulative prediction loss ($y\text{-axis}$) related to the amount of uncertain testing data included in the analysis ($x\text{-axis}$). The uncertainty effectively mitigates prediction loss for both the BNN (a) and the Gradient Boosting model (b).} 
\label{fig:perf_to_std}
\end{figure}

In the supplementary material we derive the upper and lower bounds of the binary cross-entropy loss $\mathcal{L}(y_i|\bm{x_i})$ with respect to uncertainty: 
\begin{align}
    -\log\left({\frac12 + \frac12 \sqrt{1-4\cdot Var(y_i|\bm{x_i})}}\right) \nonumber \\
    \leq \mathcal{L}(y_i|\bm{x_i})
    \leq
    -\log\left({\frac12 - \frac12 \sqrt{1-4 \cdot Var(y_i|\bm{x_i})}}\right). 
    \label{eq: deriv}
\end{align} 
These are depicted in figure \ref{fig:loss_uncertainty}. 
In the supplementary material we show how the data follows these bounds.
Areas of both low and high loss cannot be reached with high uncertainty.
In other words, when the variational posterior distribution is a good approximation of the true posterior, far-from-domain examples that cause high uncertainty will mathematically drive the loss away from the extreme values. 
From this line of thought, we hypothesize that uncertainty is able to mitigate prediction loss.
Deterministic models can output very confident predictions on inputs that lie far from their training domain, yielding high performance loss. From equation \ref{eq: deriv} we see that this is not possible for a BNN.
Additionally, the cause of a highly certain wrong prediction may be due to errors in the data, for example due to mislabelled observations.
If the variational posteriors is \textit{not} a good approximation of the true posterior, low uncertainty can still be achieved on wrong predictions, leading to bad consequences. Therefore, we evaluate empirically.

Figure \ref{fig:perf_to_std} illustrates how uncertainty relates to predictive performance. 
When the observations $\{\bm{x_1}, \dots ,\bm{x_N}\}$ are sorted according to their predictive uncertainty, it can be seen that the loss increases superlinearly for each additional data point included.
That is, the most uncertain $20\%$ of the data contribute about $20$ times the loss the most certain $20\%$ do.
In the supplementary material, we show that for the $20\%$ most certain data the area under the receiver operating characteristic curve (AUROC) approaches unity. 
Thus, low-uncertainty patients are more often classified correctly.
Therefore, by restricting classification to certain data points, we can effectively employ predictive uncertainty to calibrate the performance of the model.

A peculiar finding is depicted in figure \ref{fig: perf_std_gb}. 
We observe that uncertainty generalizes to a gradient boosting decision tree model, trained on the same data. In the case of tabular data like MIMIC, this result could prove to be particularly interesting, as neural networks are often outperformed by other non-linear models in similar tasks \citep{fernandez2014we}. 
This observation could mean that a Bayesian model can be deployed for the sole purpose of its uncertainty output, leaving the classification task to a second higher-performing model.
The total obtained cumulative loss is lower for the tree-based model, meaning that it had more confident correct predictions. This is likely due to the fact that the prediction risk for the decision tree is not mathematically bound to uncertainty. 
Note that we have not investigated when this does or does not hold. This remains worthy of further investigation. 

\subsection{Detecting Out-of-Domain Patients}
\begin{figure}
    \centering
    \begin{subfigure}{0.63\linewidth}
        \includegraphics[width=\linewidth]{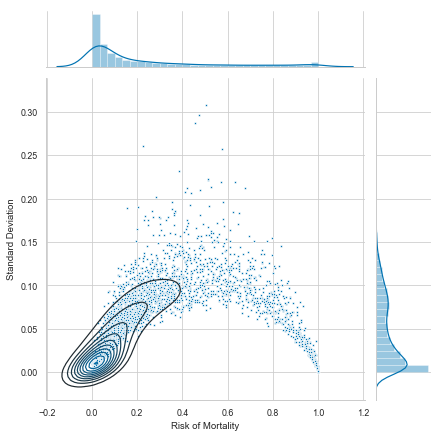}
        \caption{}
        \label{fig:pred_to_std}
    \end{subfigure}
   \begin{subfigure}{0.63\linewidth}
        \includegraphics[width=\linewidth]{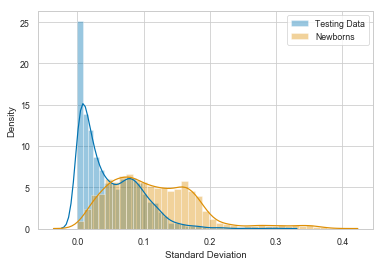}
        \caption{}
        \label{fig: train_newborns_std}
    \end{subfigure}
\caption{In plot \ref{fig:pred_to_std} we depict the predictive uncertainty ($x\text{-axis}$) related to predictive uncertainty ($y\text{-axis}$). In plot \ref{fig: train_newborns_std} we see how the BNN effectively identifies out-of-domain examples.}
\vspace{-1em}
\end{figure}
Plotting the relationship between predictions and uncertainty results in the moon-shaped scatter plot depicted in figure \ref{fig:pred_to_std}.
This corresponds to the intuition given earlier, showing that extreme predictions are only made on observations with low uncertainty.
In the central region of the graph, we see that the uncertainty has a wider spread, meaning that there are both observations with feature values occurring frequently (low uncertainty) and patients with sets of features that lie further from the previously observed domain.
Most of the mass expectedly concentrates around the surviving group of patients.
 
In the medical field, research is often conducted on a biased sample of the population of patients and therefore minorities are underrepresented \cite{baird1999new, swanson1995recruiting, giuliano2000participation, bonevski2014reaching}.
Regrettably, models and treatments are still applied to the true population, including these minorities, which can lead to adverse results.
In the MIMIC dataset, there is a group of $7,821$ newborns.
These were set aside as out-of-domain patients, and excluded from training.
We replaced obvious differences like weight and age with training data averages.
We observe in figure \ref{fig: train_newborns_std} that when the model is presented newborns, it becomes about $230\%$ more uncertain on average. 
This means that a high-uncertainty output can be used to warn a practitioner that the (combination of) symptoms and vital signs have rarely been observed before, and he or she should proceed with care. 
A plot for the newborns similar to figure \ref{fig:pred_to_std} is given in the supplementary material.
Since the characteristics of newborns greatly differ from the rest of the population, we also experimented with uncertainty on ethnic minorities.
Among others, \citet{carson1999racial} investigated the differences in responses to heart failure therapies between ethnicities. 
African American people did show different responses as a result of different features compared to the baseline. 
This motivated us to investigate model uncertainty on such an ethnic minority. 
After setting it apart, we observed that the BNN became 130\% more uncertain ($p < .0001$) on this group. 
Comparison to a deterministic baseline on these tasks is given in the supplementary material.

\section{Conclusions}
\label{sec: conclusion}
In this paper, we explored the application of Bayesian Neural Networks to improve the safety of machine-learning-based clinical decision support tools in critical areas such as the ICU.
Following the findings of \citet{blundell2015weight}, we trained a BNN on the MIMIC critical care dataset.
The objective of the model was to predict patient mortality given clinical observations and lab values.
We derived bounds on cross-entropy loss with respect to predictive uncertainty. 
Through these bounds, uncertainty is able to mitigate performance risk and loss. 
Empirically, we showed that the loss of test set predictions increased superlinearly when the patients were sorted according to their corresponding uncertainty.
Secondly, the results reveal that the uncertainty of the predictions increases significantly on out-of-domain patients. 
This suggests that in an applied setting a BNN can effectively identify patients outside of its previously observed domain.
Overall, this work demonstrates that uncertainty is effective in enhancing model trustworthiness and mitigating prediction risk and loss in an critical setting like the ICU.
A mathematical intuition as to why uncertainty generalizes to other models remains unclear.
This can be an interesting direction for future research.

\section{Acknowledgements}
\label{sec: acknowledgements}
This research was funded by Pacmed BV. We like to thank the authors of the MIMIC-III dataset for allowing us access and usage rights.

\bibliography{bib}
\bibliographystyle{icml2019}

\newpage
\begin{appendix}

\section{}
\label{sec:appa}

\begin{figure}
    \centering
    \includegraphics[width=0.7\linewidth]{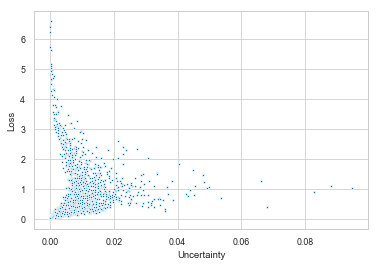}
    \caption{Illustration of how the data follows the bounds.}
    \label{fig:bounds_visualized}
\end{figure}

In this section, we show the derivation of thex  bounds on the binary cross-entropy criterion relating to predictive uncertainty. Uncertainty is computed as the variance in the predictions for each datapoint $\bm x_i$ given $T$ samples from a Bayesian posterior $p(\bm \omega| \bm X, \bm y)$. We start off with the Bhatia-Davis inequality on variance given a distribution maximum $M$ and distribution minimum $m$:
\begin{equation}
   \sigma^2 \leq (M - \mu)(\mu - m). 
\end{equation}
Solving the equality case for $\mu$,
\begin{align}
    \mu^2 - (M+m)\mu + mM + \sigma^2 = 0 \\
    \mu = \frac{M + m}{2} \pm \frac12 \sqrt{(M - m)^2 -4\sigma^2},
\end{align}
gives us bounds for the mean with respect to the variance:
\begin{align}
    \frac{M+m}{2}-\frac12 \sqrt{(M-m)^2-4\sigma^2} \leq \mu \nonumber \\ 
    \leq \frac{M + m}{2} + \frac12 \sqrt{(M-m)^2-4\sigma^2}.
    \label{eq: muineq}
\end{align}
In our case $\mu = p(y_i|\bm x_i) = \frac1T \sum_{t=1}^T p(y_{i}|f^{\omega_t}(\bm x_i))$, $\sigma^2 = Var(y_i|\bm{x_i}) = \frac1T \sum_{t=1}^T(p(y_i|{\bm x_i}) - p(y_i|f^{\bm \omega_t}(\bm{x_i}))^2$. 
$p(y_{i}|f^{\bm \omega_t}(\bm {x_i}))$ is obtained from Sigmoid activation, therefore $m = \inf_{\bm \omega}p(y_i|\bm{x_i}) = 0$ and $M = \sup_{\bm \omega}p(y_i|\bm{x_i}) = 1$.
Plugging these values and bounds in \ref{eq: muineq} into the binary cross-entropy criterion,
\begin{equation}
    \mathcal{L}(y_i|\bm{x_i}) = 
    \left.
    \begin{cases}
    \log(p(y_i|\bm{x_i})), & \text{for } y=1 \\
    \log(1 - p(y_i|\bm{x_i})), & \text{for } y=0
    \end{cases}
    \right\},
\end{equation}
 gives us a single bound (for both $y=1$ and $y=0$) on the loss w.r.t. the uncertainty:
\begin{align}
    -\log\left({\frac12 + \frac12 \sqrt{1-4\cdot Var(y_i|\bm{x_i})}}\right) \nonumber \\
    \leq \mathcal{L}(y_i|\bm{x_i})
    \leq
    -\log\left({\frac12 - \frac12 \sqrt{1-4 \cdot Var(y_i|\bm{x_i})}}\right). 
\end{align}
In figure \ref{fig:bounds_visualized} we depict how the data follows these bounds.

\section{}
\begin{figure}[H]
    \centering
    \begin{subfigure}{.7\linewidth}
        \centering
        \includegraphics[width=\linewidth]{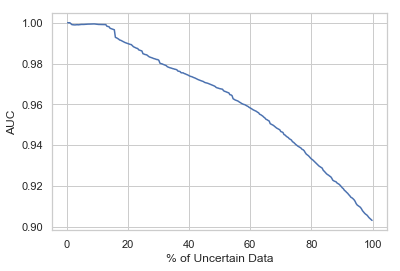}
        \caption{}
    \end{subfigure}
    \begin{subfigure}{.7\linewidth}
        \centering
        \includegraphics[width=\linewidth]{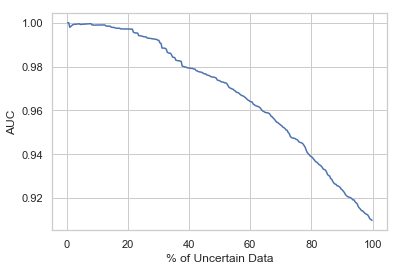}
        \caption{}
        \label{fig: perf_auc_gb}
    \end{subfigure}
\caption{Predictive performance measured in AUROC ($y\text{-axis}$) related to the amount of uncertain data included in the analysis ($x\text{-axis}$). } 
\label{fig:perf_to_auc}
\end{figure}

In figure \ref{fig:perf_to_auc} we show that the uncertainty can calibrate performance in AUC for both a BNN (a) and a Gradient Boosting model, motivating researching the usage of uncertainty in combination with readily deployed models.

\section{}
Patient data from the MIMIC-III database was used \citep{johnson2016mimic}. Vital signs, lab values and patient characteristics that were most abundantly available were gathered. Examples are blood pressure, potassium and age, respectively. To keep interpretability, we restricted ourselves to 25 clinically relevant features. Arterial and non-invasive blood pressures were combined, using the arterial blood pressure where possible. Features that lied further than 8 interquartile ranges were regarded as outliers and removed. Labels were obtained directly from the MIMIC tables, regarding expirement during the last hospital admission as a positive label. 9,237 ICU stays were set apart for testing purposes, leaving 36,944 patients for training.

\section{}
\begin{figure}[H]
    \centering
    \includegraphics[width=0.8\linewidth]{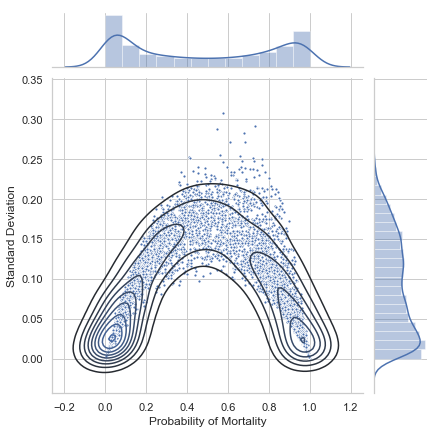}
    \caption{}
    \label{fig: pred_std_newborn}
\end{figure}

In figure \ref{fig: pred_std_newborn} we observe how the predictions on the newborns follow the same moon shape as the trained dataset. However, average uncertainty is much higher.

\section{}
Following  the approach of \citet{hendrycks2016baseline}, we compare how well sigmoid probabilities are able to detect correct classifications and out of domain patients compared to Bayesian uncertainty.

\begin{table}[H]
\caption{Comparison of error and success detection between deterministic baseline and BNN.}
\label{sample-table}
\vskip 0.15in
\begin{center}
\begin{small}
\begin{sc}
\begin{tabular}{lcccr}
\toprule
Model & AUROC & AUPR Succ & AUPR Err \\
\midrule
BNN STD     & 83.7 & 97.7 & 32.6 \\
NN Sigmoid  & 79.7 & 95.8 & 37.7 \\
\bottomrule
\end{tabular}
\end{sc}
\end{small}
\end{center}
\vskip -0.1in
\end{table}

\begin{table}[H]
\caption{Comparison of in and out of domain detection between deterministic baseline and BNN.}
\label{sample-table}
\vskip 0.15in
\begin{center}
\begin{small}
\begin{sc}
\begin{tabular}{lcccr}
\toprule
Model & AUROC & AUPR In & AUPR Out\\
\midrule
BNN STD     & 75.9 & 80.5 & 69.7 \\
NN Sigmoid  & 49.8 & 61.7 & 42.5 \\
\bottomrule
\end{tabular}
\end{sc}
\end{small}
\end{center}
\vskip -0.1in
\end{table}

\end{appendix}

\end{document}